\documentclass[journal]{IEEEtran}
\ifCLASSINFOpdf
\else
\fi

\usepackage{amsmath}
\usepackage{hyperref}
\usepackage{tcolorbox}
\usepackage{booktabs}
\usepackage{enumitem}
\begin{document}
%
\title{Semantics of the Black-Box: Can knowledge graphs help make deep learning systems more interpretable and explainable?}
%
%
%

\author{Manas~Gaur,\IEEEmembership{}
        Keyur~Faldu,\IEEEmembership{}
        Amit~Sheth\IEEEmembership{}
\thanks{Manas Gaur is a PhD student at the Artificial Intelligence Institute, University of South Carolina, e-mail: mgaur@sc.edu}
\thanks{Keyur Faldu is Chief Data Scientist at Embibe, e-mail: k@embibe.com}
\thanks{Amit Sheth is the Founding Director at the Artificial Intelligence Institute, University of South Carolina, e-mail: amit@sc.edu}
\thanks{Manuscript accepted to IEEE Internet Computing, October 30, 2020}}

%
%

\markboth{}{}
%



\maketitle

\begin{abstract}
The recent series of innovations in deep learning (DL) have shown enormous potential to impact individuals and society, both positively and negatively. The DL models utilizing massive computing power and enormous datasets have significantly outperformed prior historical benchmarks on increasingly difficult, well-defined research tasks across technology domains such as computer vision, natural language processing, signal processing, and human-computer interactions. However, the Black-Box nature of DL models and their over-reliance on massive amounts of data condensed into labels and dense representations poses challenges for interpretability and explainability of the system. Furthermore, DLs have not yet been proven in their ability to effectively utilize relevant domain knowledge and experience critical to human understanding. This aspect is missing in early data-focused approaches and necessitated knowledge-infused learning and other strategies to incorporate computational knowledge. This article demonstrates how knowledge, provided as a knowledge graph, is incorporated into DL methods using knowledge-infused learning, which is one of the strategies. We then discuss how this makes a fundamental difference in the interpretability and explainability of current approaches, and illustrate it with examples from natural language processing for healthcare and education applications.

\end{abstract}

\begin{IEEEkeywords}
Knowledge Graphs, Knowledge Infusion, Neuro-Symbolic AI, Explainability, Interpretability, Black-Box Deep Learning, Mental Healthcare, Education Technology
\end{IEEEkeywords}

%
\IEEEpeerreviewmaketitle

\section{Black-Box Nature of DL Models }
%
%
%
%
\IEEEPARstart{DL}{models} are complex and opaque. The cascading sequences of linear and non-linear mathematical transformations learned by models comprising millions of parameters are beyond human comprehension and reasoning. This renders them as “Black-Box” models for decision-making. 
For example, in a trivial case of question answering: \textit{Context:} I sometimes wonder how many alcoholics are relapsing under the lockdowns; \textit{Question:} Does the person have an addiction? \textit{Response generated from a pre-trained seq2seq model:}  Yes; 

\noindent Comparing this to a non-trivial case of question answering: \textit{Context:} Then others that insisted that what I have is depression even though manic episodes aren't characteristic to depression.   I dread having to retread all this again because the clinic where I get my mental health addressed is closing down due to loss in business caused by the pandemic; 
\textit{Question:} Does the person suffer from depression?; \textit{Response generated from a pre-trained seq2seq model:} Yes (the correct answer is no).

In such scenarios, it is exceptionally challenging to probe the model's mechanism without the support of background knowledge \cite{camburu2018snli}. Although Neural Attention Models (NAM) \cite{gaur2019shades} are endowed with a certain degree of interpretability in visualizing parts of the sentence which are focused on answering the question, they cannot provide further explanations to answer the question in a human-understandable format. Furthermore, recent research has highlighted challenges concerning the model behavior in question answering domains, particularly towards queries with disjunctive clauses (questions that contain or) \cite{ren2020query2box}.  We have observed unpredictable responses to questions like Are you feeling nervous or anxious or on edge?  Is the feeling of restlessness due to stress or anxiety? Does an employee own a company or work for a company? 

The ability to fine-tune pre-trained models trained in unsupervised settings to a downstream task has alleviated the need for an extensive labeled dataset. This introduces the challenge of explaining decisions using insights from limited labeled data. Research practices have often resorted to using frozen datasets, intending to surpass benchmark test-data-accuracy. As a result, DL models have outperformed human baselines on specific datasets in terms of ‘test-data-accuracy’ but still fail to generalize intuitive cases in real-world scenarios. Ribeiro et al.  introduced CHECKLIST, a task agnostic framework to test the generalisability of NLP models, and found that these models are vulnerable to test data prepared with linguistically intuitive rules \cite{ribeiro2020beyond}. The bottom line is that most state-of-the-art DL approaches are not integrated with prior knowledge, a necessary condition for explaining the predictions, and interpreting the model mechanism for appropriate probing.


\section{Need for Explainability and Interpretability}
The Black-Box nature of DL models needs to be addressed to foster trust among users and domain experts to be used with higher confidence. This can also facilitate broader assimilation in a variety of domains. In healthcare, clinicians routinely choose methods that allow them to understand how an outcome was derived compared to an objectively superior method that cannot be explained. In education, tracing students' learning outcomes with attribution to weak academic and behavioral areas is a better tool for teachers compared with the ability to predict only a student's performance. As a result, explainability and interpretability for the DL models have focused on recent key research areas. 
\begin{tcolorbox}
Prior research has attempted to establish a generic definition of explainability, which is the ability to generate human-comprehensible explanations around the decision-making process \cite{chari2020explanation}\cite{oltramari2020neuro}\cite{samek2019towards}. In contrast, interpretability is the ability to discern the internal mechanisms of any module. Key reasons we need explainability and interpretability are to: 
\begin{enumerate}[label=(\alph*)]
    \item Trace and verify the fine-grained supporting information in safety-critical systems (e.g., autonomous driving vehicles)
    \item  Support evolving events and discern necessary context - as the underlying facts (data) may not be static (e.g. newer findings replace earlier findings), and meta-information such as time and space/location may be critical in understanding, interpreting, and explaining results (e.g., identifying emerging sub-events in natural disasters \cite{arachie2020unsupervised})
    \item Discover inherent bias in the model’s predictive strategy (e.g., contextual modeling \cite{gaur2019shades}\cite{gaur2020knowledge})
    \item Prevent prediction errors in unintuitive scenarios (e.g. adversarial examples \cite{gilpin2018explaining}, CHECKLIST \cite{ribeiro2020beyond})
    \item Make sure minor perturbations  in inputs are handled in robust ways
    \item After enhancement of training data through knowledge. graph (KG), proactive detection of outliers is possible.
    \item Gather new knowledge insights to further enable research and  ensure that acquired domain knowledge is getting leveraged for the decision making process.
\end{enumerate}
The enumerated list combines the need for ``explainability and interpretability'' as it applies to DL systems and enhancements obtained by integrating a knowledge graph-based approach. Addressing these concerns would foster confidence among domain experts and trust among end-users. The terms explainability and interpretability are often used interchangeably in the prior research without clear distinctions and the different roles that they play \cite{gilpin2018explaining}\cite{mittelstadt2019explaining}\cite{biran2017explanation}.
\end{tcolorbox}
\section{Defining Explainability and Interpretability with Knowledge-infusion}
Making explanations on the model behavior is subjective to the problem from the stakeholders. A set of privileged knowledge (e.g., domain expertise, advice specific to the situation) must be infused to comprehend the model outcomes and interpret its functioning. The methods for infusing knowledge in DL models occurs through a set of neuro-symbolic procedures that enrich the dataset with concepts and relationships from multiple KGs or ontologies, which assists end-users in decision making. For instance: Taxonomic Knowledge (e.g., Columbia Suicide Severity Risk Scale (C-SSRS)), Relational Knowledge (e.g., ConceptNet), Metaphorical Knowledge, and Behavioral Knowledge (e.g., LIWC) are necessary forms of external contextual information required to understand online conversations, mainly when the problem is a low resource (insufficient benchmark datasets and unlabeled corpus for transfer learning) \cite{purohit2020knowledge}. Also, knowledge-infusion during the model learning using information-theoretic loss function (e.g., KL divergence \cite{gaur2019shades}) can check conceptual drifting at the representational level through weak-supervision from KG. Alternatively, the loss of information during learning can be supplemented by augmenting the abstract information from KG to layered representation in DL models through various mathematical operations (e.g., pointwise multiplication, concatenation). 
Fusing the relevant information from KG to hidden representations in DL allows quantitative and qualitative assessment of its functioning, which we define as knowledge-infused interpretability. One possible way to do knowledge infusion for discerning model behavior is through an architecture having layers of DL connected to KG through a function. Such a function can be a “knowledge-aware loss function” that computes the loss in information at each layer per epoch. Or a function can be “knowledge-aware propagation function,” which computes the loss information and transfers missing information through mathematical operations. The computing of loss in information is performed by tracing the KG through the embedding of hidden layers and printing the concepts and relationships that the model is learning through pattern recognition. On the other hand, having a ground-truth subgraph of KG would help in calculating the distance between the concepts learned by hidden layers and actual concepts in the subgraph (all done at the embedding level), which can be used to modulate the hidden embeddings and thus leading to faster convergence with model interpretations.
In achieving interpretability and explainability through knowledge-infusion, these two terminologies can be differentiated as Explainability would cater to why the prediction has been made. In contrast, interpretability would unfold the ability to understand patterns learned or knowledge acquired by the system. Any explainable system has to be interpretable, but the reverse is not valid. Explainability can be evaluated with its interpretability and completeness.

\textbf{“Empirically, an explainable system would comprise of collectively exhaustive interpretable subsystems and orchestration among them. More often than not, explanations would be in natural language explaining the decision, while interpretations can be statistical or conceptual (using either generic or domain-specific KG \cite{bhatt2018enhancing,gaur2020knowledge}) in nature pertaining to its inner functioning.”} 

Explanations can be thought of as answers to cascading “why questions” in the context of model prediction until there is no longer a need to ask another “why question”. Explanations are expected to be faithful and plausible. Faithfulness defines how well the explanation correlates with the model's decision making. It is considered plausible when it has a human-understandable justification for the prediction \cite{faldu2018adaptive}\cite{goel2015nature}. Such systems are considered to be potentially useful for real-world decision making in various domains, especially healthcare and education technology. 
\begin{table*}[ht]
\centering
\begin{tabular}{p{2.5cm}p{3.5cm}p{5cm}p{5cm}} 
\toprule[2.5pt]
\textbf{Focus Area} & \textbf{Approach} & \textbf{Methods} & \textbf{Interpretable outcomes}\\ \midrule[1pt] 
Input Features & Proxy Functions & LIME, Gradients, Saliency maps, SmoothGrad & Linear Model Coefficients, Relative Features Importance\\ 
 Input Features & Occlusion or Perturbations & SHAP, Integrated Gradients, DeepLift, PDA, Activation Maximization & Relative Features Importance\\
 Input Features & Tracing Positive Contributions &  LRP, Deconvolutions, Guided Backpropagation & Relative Features Importance\\ 
Representations & Projections & PCA, ICA, NMF, Conicity & Interpretable Vector Representations Space\\
Attentions & Transformations & Effective attention, attention flow & Interpretable Attentions Distribution\\
Neurons, Layers Contribution & Transformations & Conductance, Layer Relevance Propagation, DIFFPOOL & Neurons, Layers Contribution\\
Encoded Knowledge in Representations, Attentions, Layers  & Probes & Knowledge Probes, Diagnostic Classifiers, Auxiliary Tasks & Interpretable Vector Representations Space, Attentions Distribution, Relative Feature Importance\\
\bottomrule[2.5pt]
\end{tabular}
\vspace{1em}
\caption{\label{tab:comparison}Prior approaches and methods to discover interpretable subsystems. Though it forms the superset of the taxonomy outlined in Gilpin et al., these methods are not sufficient to achieve explainability and interpretability of the level required for decision making in high stakes problems, including  mental healthcare and education technology.}
\end{table*}

\section{Limitations of Learning using Distributional Semantics }
The recent success of DL language models in NLP has been attributed to self-supervised objectives over a large volume of unlabeled data such as BERT, RoBERTa, and T5. These models learn distributional semantics about the text and relationships between phrases. It is also an active research area to probe whether these models learn linguistic knowledge like part of speech and dependency tree. Many probing experiments have given insights into linguistic patterns discovered by different layers and internal components like attention heads of neural networks. However, it is still an open question, how much semantic knowledge is learned by these models, specifically when the semantics are not expressed with statistically significant patterns in the data. This becomes evident when these models fail to learn facts around concepts.  

Furthermore, there is a growing trend of fine-tuning the pre-trained models with limited labeled data. These have given success when (a) the distribution of the labeled dataset is similar to unlabeled data used for pre-training, (b) tasks are relatively  straightforward like natural language entailment and span extractive question answering. However, real-world scenarios are often more complex, which poses the following challenges: (a) Fine-tuning such models for domain-specific tasks with limited labeled data may not be sufficient to learn domain knowledge  as these data may not be able to capture domain knowledge. (b) Similarly, self-supervised training objectives over unlabeled data is not attempting to learn the domain knowledge required for real-world adoption. (c) Generating explanations and interpretations would be even more challenging as internal layers or mechanisms of distributional semantics would not represent specific domain knowledge. 

\section{Knowledge Graph Infusion for Better Explainability}
\begin{figure*}[!htbp]
    \centering
    \includegraphics[width = \textwidth]{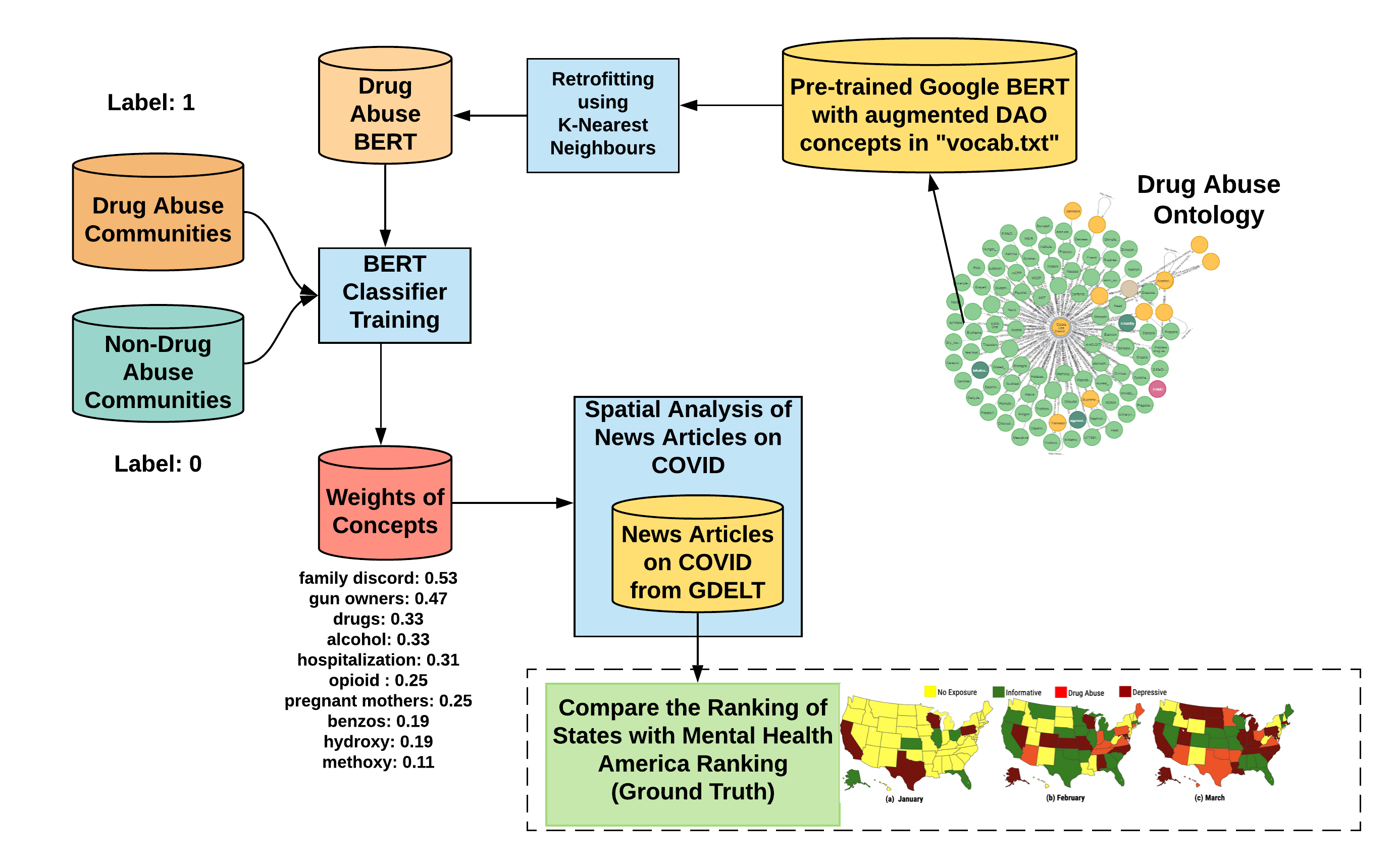}
    \caption{Illustration of shallow infusion wherein the Google BERT model is reconstructed after concepts from Drug Abuse Ontology (DAO) were appended to existing vocabulary to assess the negative media exposure during COVID-19. This weakly supervised approach with knowledge infusion is practical when it is difficult to annotate millions of news articles and procure labeled dataset for depression and drug abuse}
    \label{fig:fig 1}
\end{figure*}

Aforementioned concerns in distributional semantics calls for a specific need to infuse domain knowledge in deep neural models for better predictive performance and the descriptive performance of generating explanations and interpretations. Infusing domain knowledge in DL models can be categorized into the shallow infusion, semi-deep infusion, and deep infusion \cite{gaur2019shades}. In shallow infusion, both the external information and method of knowledge infusion is shallow, i.e., Word2Vec or GloVE embeddings are reconstructed using domain knowledge as features (see Figure \ref{fig:fig 1} \cite{alambo2020depressive}). On the other hand, deep infusion of knowledge is a paradigm that couples the latent representation learned by deep neural networks with the KGs exploiting the semantic relationships between entities \cite{kursuncu2019knowledge}. 

Recently, there has been a flurry of research to experiment with different techniques to infuse knowledge graphs or domain rules to deep neural models for different needs. Mainly they can be categorized as follows: (a) Fusion of relevant external knowledge as an additional context in the query or as external knowledge-based features \cite{kursuncu2019modeling},\cite{yao2019kg}, (b) Infusing KG embedding to hidden layers of neural models for explainable decision making \cite{wang2019explainable}, (c) Infusing contextual representations from relevant subgraph or paths of KG through either concatenation or pooling or non-linear transformations \cite{yang2019leveraging}, (d) Leveraging KG to alter attention mechanism and pre-training of DL model \cite{choi2019graph}, and (e) Developing approaches that uses  domain-specific KG using Integer Linear Programming for non-trivial natural language inference \cite{khashabi2016question}. 

\section{Explanations and Interpretations for Healthcare Domain Use Cases}
\begin{figure*}[!htbp]
    \centering
    \includegraphics[width = \textwidth]{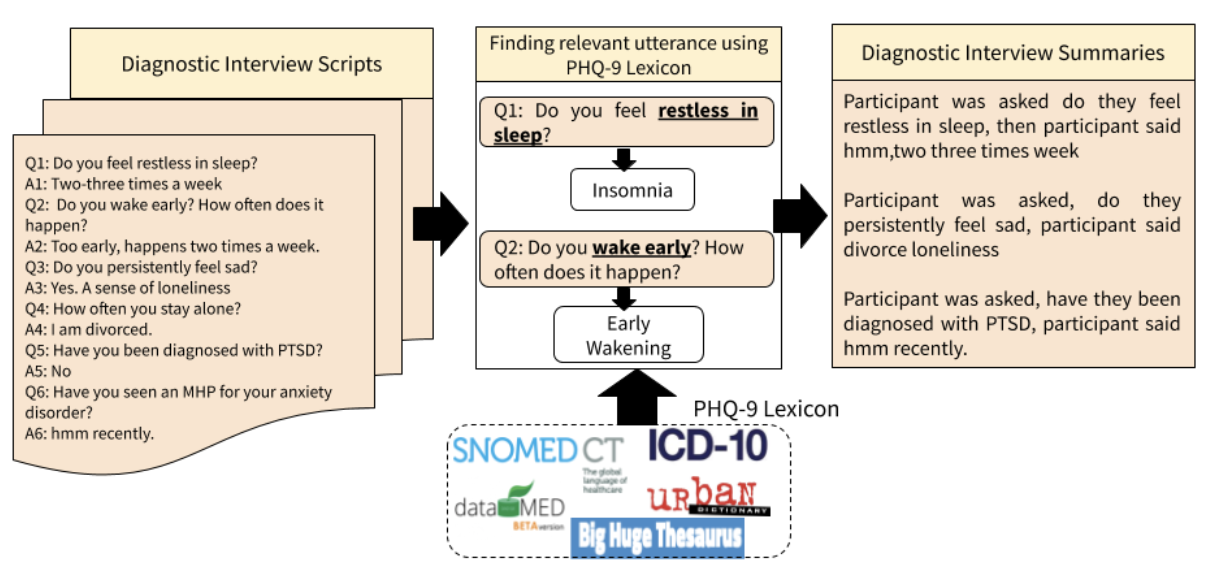}
    \caption{ Overview of Knowledge-infused Abstractive Summarization (KiAS) for an interview snippet of a patient's responses to the question asked by Ellie (virtual interviewer). Phrases relevant to mental health are identified using the PHQ-9 lexicon. The contextual similarity between utterances is calculated through a retrofitted embedding model. The resulting summaries contain relevant questions and meaningful responses. KiAS strategy recognizes the semantic similarity between Anxiety disorder [SNOMEDCT ID: 197480006] and PTSD [SNOMEDCT ID: 47505003] as described in the SNOMED-CT hierarchy through an "is-a" relationship and recorded in the PHQ-9 lexicon. It associates responses of anxiety and PTSD interchangeably}
    \label{fig:fig 2}
\end{figure*}

Consider an application domain of summarization, which is at the intersection of natural language understanding and DL. An extensive body of research has investigated summarizing news articles, meeting logs, and financial/legal contracts that are often templatized. On the other hand, patient-clinician conversations are open-ended as the clinician tries to reflect on the patient's responses. Consequently, the interview forms an inherent structure, that raises challenges in natural language understanding: (1) Anaphora- where sentences are purposefully paraphrased to elicit meaningful responses from an agent (or user); (2) The clinical conversation contains implicit references to healthcare conditions, developing sparsity in the clinically-relevant concepts. 
Such a problem scenario requires a model to capture the context of the conversation (see Figure \ref{fig:fig 2}). 

\begin{figure*}[ht]
    \centering
    \includegraphics[width = \textwidth]{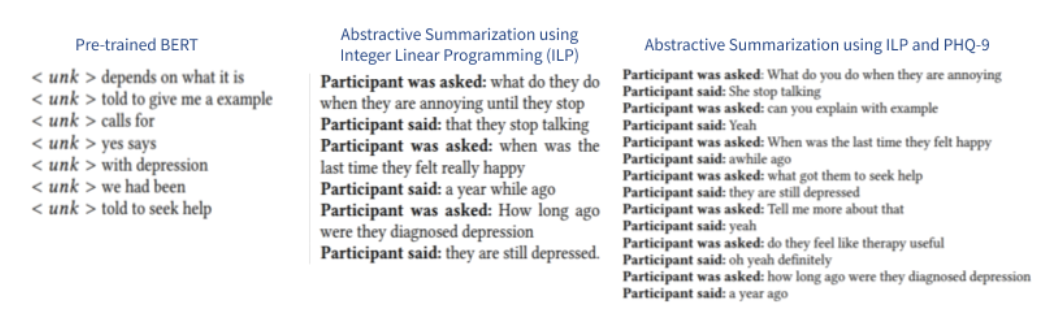}
    \caption{In an experiment to summarize a transcript of a 12 minute recorded conversation between a patient and clinician (https://bit.ly/patid313), ILP with PHQ-9 knowledge could align appropriate responses to the question much better than simple ILP based abstractive summarization and pre-trained BERT. KiAS generates summaries (7 sentences on an average) that capture informative questions and responses exchanged during long (58 sentences on an average), ambiguous, and sparse clinical diagnostic interviews}
    \label{fig:fig 3}
\end{figure*}

The summarization of clinical conversation requires meaningful responses to be associated with clinically-relevant questions while resolving anaphora problem. State-of-the-art language models (e.g., BERT) trained over the large-scale corpus of news articles fail to capture the question's context and assess the importance of a response from the clinician perspective. In addition, BERT-like models' fine-tuning is not helpful because updating the model parameter needs to be governed by a structure that can abstractly describe a clinical conversation with stratified knowledge (see Figure \ref{fig:fig 3}). Hence, recently, Integer Linear Programming (ILP)-based summarization approaches have gained sufficient traction from the NLP community\footnote{\url{https://www.jmir.org/preprint/20865}}. The optimization framework is interpretable as knowledge is incorporated as constraints. The outcome is explainable because optimization criteria reflect on the end-users requirement.
\begin{figure*}[!htbp]
    \centering
    \includegraphics[width = \textwidth]{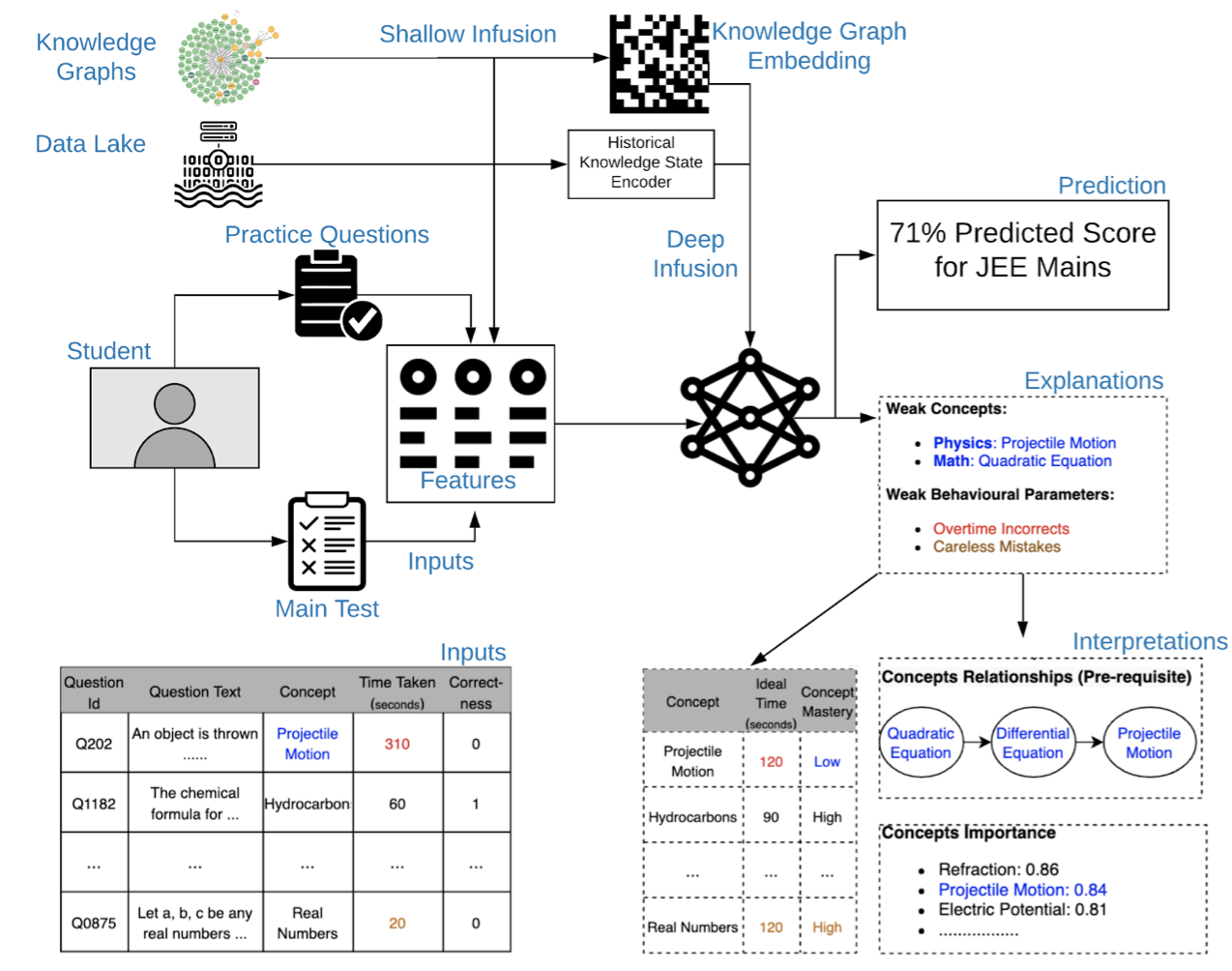}
    \caption{Interpretations and Explanations for Students Performance Prediction. Student’s attempt stream in practice or test session is fed as ``Inputs''. Domain knowledge present in knowledge graphs and data lakes can be “shallow infused” in features, or “deep infused” in the model. Explanations and Interpretations are generated along with prediction of student score.}
    \label{fig:fig 4}
\end{figure*}
\section{Explanations and Interpretations for Education Domain Use Cases}
Let’s take an example of a use case in the education domain (see Figure 4). We illustrate this figure with respect to two types of knowledge infusion - shallow and deep infusion. Shallow infusion covers the scenario of using knowledge at the start of the DL pipeline and deep infusion refers to using knowledge at every layer of the pipeline \footnote{\url{http://bit.ly/kidl2020}}.  Student’s score prediction is one of the key problems to assess the student’s true potential for a goal. In the context of this problem, explanations of this prediction are really important as they reveal what governs the student score, both strengths, and weaknesses on the academic and behavioral aspects. These explanations can potentially induce remedial actions from the student and mentor. 

The domain knowledge for student performance prediction can be thought of as (a) Academic knowledge graph, set of concepts, it's metadata and relations between concepts that plays an important role in tracing student's concepts mastery \cite{dhavala2020auto}\cite{kursuncu2019knowledge}, and (b) Engagement and learning patterns from historical student’s data \cite{faldu2018adaptive}. Students' historical knowledge state could be derived from this data \cite{donda2020framework}. 

The domain knowledge infused model would be able to explain student's performance by tracing the predicted score to weak concepts, furthermore, to the cause of these weak concepts. Infusing academic knowledge graphs and a student's historical knowledge state makes it possible to trace academic weakness until the deepest pre-requisite concept impacts student score \cite{wang2019deep}. As shown in Figure \ref{fig:fig 4}, the student has got an attempt wrong on a question of concept ``Projectile Motion'', using a knowledge graph and student’s knowledge state it could be traced deeper to the previous grade concept ``Quadratic Equation''.

Students’ behavior is an equally important part of tracing learning outcomes \cite{donda2020framework}. Assessments or diagnostic tests are also created using such behavioral profiling \cite{dhavala2020auto}. Student's tendency to answer a question without comprehending or thinking through can lead to “Careless Mistakes”, which could be the result of guessing the answer or expressing over-confidence. Similarly, the inability to leave a question unanswered to focus more on answering other questions could be termed as a lack of decisiveness. Which could lead to ``Overtime Incorrects'' attempts.  Domain knowledge is needed to classify attempts to a ``Careless Mistakes'', or an  ``Overtime Incorrects'' attempt. Figure 4 shows an example of generating explanations with student’s behavioral traits like ``Careless Mistakes'', ``Overtime Incorrects'' by infusing domain knowledge into the model. 

Interpretations around these explanations could be traced back to the student's attempts-stream as input, domain knowledge, and data lake to understand how these are derived by applying interpretability techniques mentioned in Table 1. Moreover, the impact of such explanations on student’s performance prediction could also be derived and it could furnish as a foundation for recommendations and estimating learning outcomes upon student’s action for the same \cite{donda2020framework}.

\section{Discussion}
In this article, we highlighted the need for explanations and interpretations for the domain adoption of AI models, particularly in healthcare and education technology. We overviewed existing statistical methods and metrics devised to quantitatively assess the explainability of the model and interpretability of its mechanisms. Existing frameworks categorized as post-hoc Interpretability, counterfactual explanations, and rule-based explanations fall short in providing answers to the following open questions: 
\begin{enumerate}
    \item Can the model mine (varied) relationships from the existing text?
    \item Can the model reliably classify entities into known ontology?
    \item Can the model answer the question with trust and transparency?
    \item Is it possible to measure the model's ``reasonability'' and ``meaningfulness'' of the response to a question?
    \item How much context is needed for the model to provide a precise response?
\end{enumerate} 
An emerging trend to fine-tune a pre-trained model on limited labeled data for a downstream task and the inability of distributional semantics learning to capture domain-specific knowledge pose limitations in addressing the above questions. We noticed the necessity of KG as an integral component in neuro-symbolic AI systems with capabilities to generate explainable outcomes and interpretability through tracing over KG. Future advances in this area would seek AI systems that can help stakeholders (e.g., instructors, public health experts) to conceptually understand the working of involved AI systems \cite{huk2018multimodal}. There is also a pressing requirement for benchmark datasets which assess the quality of explanations derived from model outcomes and interpretability of the algorithms in achieving explanations \cite{ribeiro2020beyond} \cite{rudin2019stop}.

\bibliographystyle{IEEEtran}
\bibliography{references.bib}

\end{document}